\documentclass[sigconf]{acmart}

\usepackage{booktabs}
\usepackage{multirow}
\usepackage{graphicx}
\usepackage{subcaption}
\usepackage{amsmath}
\usepackage{xspace}
\usepackage{url}
\usepackage{hyperref}
\usepackage{cleveref}

\acmDOI{10.1145/XXXXXXX.XXXXXXX}
\acmISBN{978-x-xxxx-xxxx-x/XX/XX}
\acmConference[CONF'XX]{TODO: Full Venue Name}{TODO: Month Year}{TODO: City, Country}
\acmYear{2026}
\copyrightyear{2026}
\acmPrice{15.00}

\newcommand{\fase}{\textsc{fase}\xspace}
\newcommand{\stgnn}{\textsc{stgnn}\xspace}
\newcommand{\zinb}{ZINB\xspace}

\newcommand{\eg}{\emph{e.g.},\xspace}

\begin{document}

\title{\fase: A Fairness-Aware Spatiotemporal Event Graph Framework for Predictive Policing}


\author{Pronob Kumar Barman}
\affiliation{\institution{Department of Information Systems \\ University of Maryland, Baltimore County} \city{Baltimore} \country{MD, USA}}

\author{Pronoy Kumar Barman}
\affiliation{\institution{Department of Statistics \\ Jagannath University} \city{Dhaka} \country{Bangladesh}}

\author{Plaban Kumar Barman}
\affiliation{\institution{MBBS \\ Shaheed M. Monsur Ali Medical College} \city{Sirajganj} \country{Bangladesh}}

\author{Rohan Mandar Salvi}
\affiliation{\institution{Department of Information Systems \\ University of Maryland, Baltimore County} \city{Baltimore} \country{MD, USA}}

\begin{abstract}
Predictive policing systems that allocate patrol resources purely on
predicted crime risk risk perpetuating and amplifying racial disparities
through feedback-driven data bias.
We present \fase\ (\textbf{F}airness-\textbf{A}ware \textbf{S}patiotemporal
\textbf{E}vent Graph), a five-phase pipeline that couples spatiotemporal
crime prediction with fairness-constrained patrol allocation and a
closed-loop deployment feedback simulator.

\fase\ represents Baltimore, Maryland as a graph of 25 ZIP Code
Tabulation Areas (ZCTAs) and models 139,982 Part-1 crime incidents
over 2017--2019 at one-hour temporal resolution, yielding a
$25 \times 26{,}280 \times 13$ feature tensor with 83.21\% sparsity.
The prediction engine fuses a Graph WaveNet-style Spatiotemporal Graph
Neural Network (\stgnn) for capturing spatial diffusion and periodicity
with a GPU-vectorised multivariate Hawkes excitation layer for modelling
self-exciting temporal clustering.
Predictions are decoded by a Zero-Inflated Negative Binomial (\zinb)
head---an appropriate distributional choice for the overdispersed,
zero-heavy count structure of hourly crime data.
The best model achieves a validation \zinb\ loss of 0.4800 at epoch 91
and a held-out test loss of 0.4857.

Patrol units are allocated by a fairness-constrained linear programme
that maximises risk-weighted coverage subject to a Demographic Impact
Ratio (DIR) constraint $|\mathrm{DIR}-1| \le 0.05$,
where DIR compares mean patrol intensity in minority versus
non-minority ZCTAs.
Over six simulated deployment cycles the allocation DIR remains
within $[0.9928, 1.0262]$, while risk-weighted coverage ranges
from 0.876 to 0.936.
A persistent detection-rate gap between minority and non-minority ZCTAs
(approximately 3.5 percentage points across all cycles) highlights that
allocation-level fairness constraints do not fully eliminate feedback-induced
observational bias in the retraining data, a limitation we analyse explicitly.

\end{abstract}



\maketitle

\section{Introduction}
\label{sec:introduction}

Predictive policing---the use of algorithmic methods to forecast where
and when crime is likely to occur---has attracted sustained attention
from both the machine learning community and public-policy researchers.
Systems informed by historical incident data have been deployed in dozens
of major U.S. cities~\cite{mohler2015randomized,ferguson2017policing}.
Proponents argue that data-driven resource allocation can reduce crime by
concentrating patrol effort where it is most needed~\cite{mohler2015randomized}.
Critics, however, have documented that systems trained on historical
arrest and incident data encode pre-existing racial and geographic
biases~\cite{richardson2019dirty,lum2016predict,berk2021fairness,chouldechova2017fair,dressel2018accuracy},
and that the act of increased patrol itself generates more detected
incidents in surveilled areas---a feedback loop that progressively
concentrates enforcement in communities that are already
over-policed~\cite{ensign2018runaway,barman2025unmasking}.

The algorithmic fairness literature has produced a rich body of work on
measuring and mitigating bias in predictive
systems~\cite{mehrabi2021survey,selbst2019fairness,hung2023predictive,hernandez2026quantifying,ziosi2024evidence}.
Reviews focused specifically on predictive policing reveal structural
tensions between accuracy optimisation and equitable
outcomes~\cite{alikhademi2022review,almasoud2024algorithmic}.
Counterfactual and causal approaches to fairness offer principled
alternatives to distributional constraints, though they remain
computationally demanding at deployment
scale~\cite{kim2021counterfactual,zhang2018fairness}.
Recent comparative simulation studies have further shown that bias
metrics are city-specific and year-variant, and cannot be assumed stable
across deployment cycles~\cite{semsar2026comparative,barman2025unmasking}.

Existing work has addressed pieces of the prediction-allocation-feedback
problem in isolation.
Graph-based spatiotemporal forecasting methods such as Graph
WaveNet~\cite{wu2019graphwavenet}, DCRNN~\cite{li2018dcrnn}, and
spatiotemporal multi-graph networks~\cite{wang2024stmgnn} deliver strong
predictive performance but do not incorporate fairness constraints.
Recent work on fair crime prediction highlights the additional challenge
that historical under-reporting skews training
data~\cite{wu2024fairness,wang2023pursuit}.
Fairness-aware resource allocation provides principled LP
formulations~\cite{wang2024police} but typically assumes exogenous,
fixed predictions.
Work on feedback and measurement
bias~\cite{ensign2018runaway,selbst2019fairness} identifies the problem
structurally but does not provide a closed-loop simulator that quantifies
feedback amplification cycle by cycle.

\medskip
\noindent\textbf{Our Contributions.}
Building on our prior multi-city GAN simulation
framework~\cite{barman2025unmasking}, this paper presents \fase\
(\textbf{F}airness-\textbf{A}ware \textbf{S}patiotemporal \textbf{E}vent
Graph), a unified, end-to-end pipeline that addresses prediction,
allocation, and feedback bias simultaneously.  Specifically:

\begin{enumerate}
  \item \textbf{Dual-engine spatiotemporal predictor.}
    We combine a Graph WaveNet-style \stgnn\ with a GPU-vec\-torised
    multivariate Hawkes excitation layer.  The \stgnn\ captures
    long-range spatial spillover and periodic temporal patterns via
    dilated causal convolutions and graph diffusion; the Hawkes layer
    models self-exciting, history-dependent event clustering.
    Outputs are decoded by a \zinb\ head---appropriate for the 83.21\%
    sparse, overdispersed hourly count data.

  \item \textbf{Fairness-constrained allocation.}
    A linear programme allocates $B = 60$ patrol units per timestep
    across 25 ZCTAs, maximising risk-weighted coverage subject to
    $|\mathrm{DIR}-1|\le\varepsilon$ with $\varepsilon=0.05$.
    The constraint is hard-coded in the optimisation and is verified
    empirically over every deployment cycle.

  \item \textbf{Closed-loop deployment feedback simulator.}
    We implement a $K=6$ cycle simulation in which patrol allocations
    modulate detection probability, producing observed (patrol-biased)
    counts that are fed back into model retraining.
    The simulator uses a deterministic expected-value detection model to
    isolate the fairness signal from stochastic noise.
    We track DIR, Gini concentration, risk-weighted coverage, and
    group-specific detection rates across all cycles.

  \item \textbf{Open-source reproducibility.}
    All code, SLURM job scripts, training logs, and simulation results
    are publicly available in the artifact repository (see
    \Cref{sec:artifact}).
    Every number reported in this paper is traceable to a logged artefact
    in the repository.
\end{enumerate}

\medskip
\noindent\textbf{Scope and Caveats.}
\fase\ is a research prototype evaluated on a single city (Baltimore, MD)
over a three-year historical window.
We do not claim real-world deployment readiness.
Empirical results are presented without comparison to baseline models,
as no baselines have been run under the same data and evaluation protocol;
we discuss this limitation in \Cref{sec:limitations}.
The allocation fairness metric (DIR) addresses patrol-intensity parity
but does not capture all relevant dimensions of algorithmic
fairness~\cite{berk2021fairness,mehrabi2021survey}.

The remainder of this paper is organised as follows.
\Cref{sec:method} describes the five-phase \fase\ architecture.
\Cref{sec:setup} details the experimental configuration.
\Cref{sec:results} presents all empirical results.
\Cref{sec:limitations} discusses limitations.
\Cref{sec:conclusion} concludes.

\section{\fase\ Framework}
\label{sec:method}

\begin{figure*}[t]
  \centering
  \includegraphics[width=\linewidth]{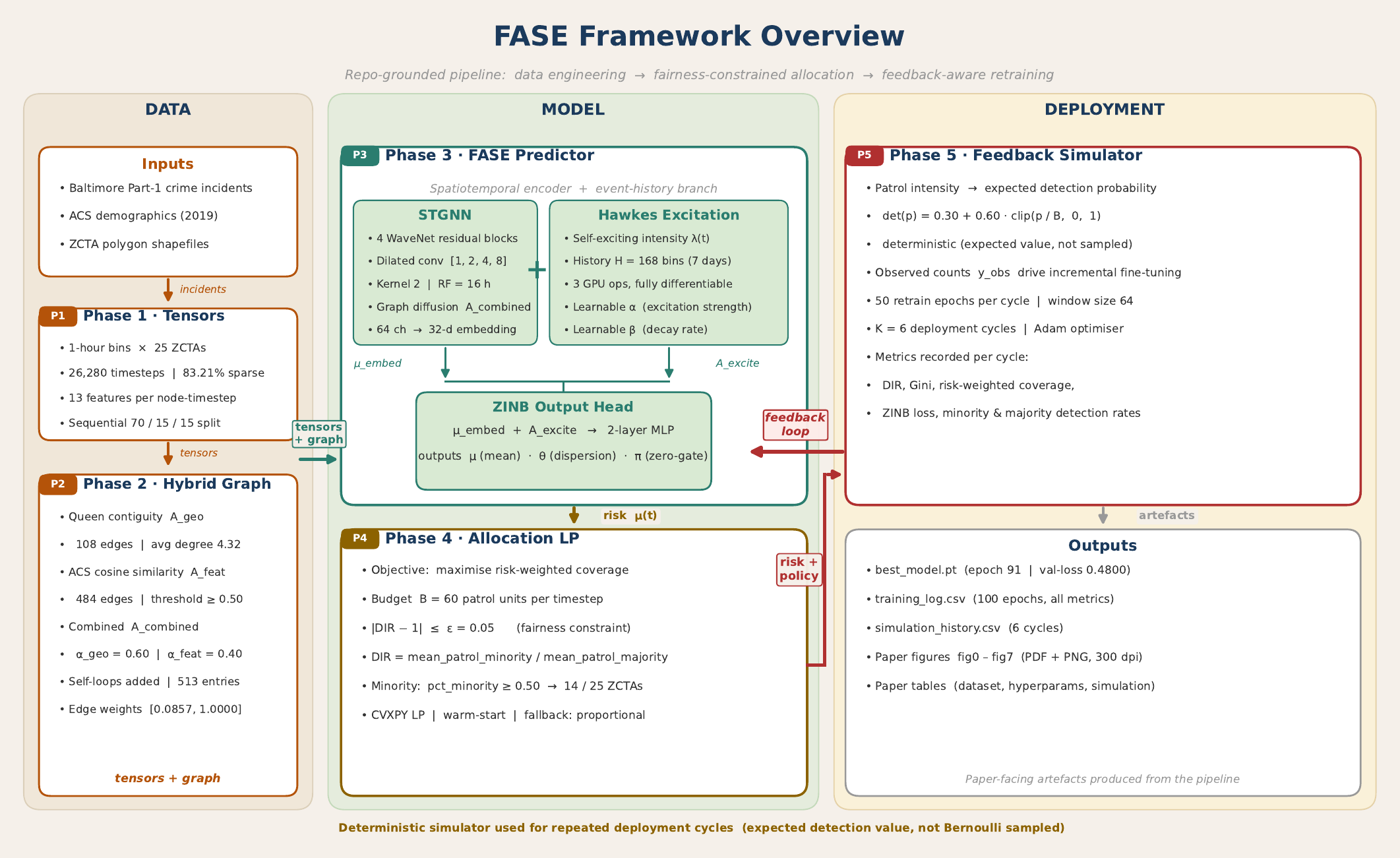}
  \caption{\fase\ pipeline overview. Arrows show data flow; the dashed
    feedback loop (Phase~5 $\to$ Phase~3) illustrates how observed,
    patrol-biased counts are fed back into incremental model retraining.}
  \label{fig:framework}
\end{figure*}

\Cref{fig:framework} gives an overview of the \fase\ pipeline.
The system consists of five sequential phases: data engineering and tensor
construction (Phase~1), hybrid graph construction (Phase~2), dual-engine
spatiotemporal prediction (Phase~3), fairness-constrained patrol allocation
(Phase~4), and closed-loop deployment feedback simulation (Phase~5).

\subsection{Phase 1: Data Engineering and Tensor Construction}
\label{sec:phase1}

\textbf{Data source.}
We use Baltimore Part-1 crime incidents for 2017--2019~\cite{baltimore2019data},
clipped to a WGS-84 bounding box covering latitudes 39.197--39.372 and
longitudes $-$76.713--$-$76.529.
After parsing and coordinate filtering, 145,178 incidents are loaded.

\textbf{Spatial join.}
Incidents are assigned to Census ZIP Code Tabulation Area (ZCTA) polygons
via a spatial inner join.  Of the 145,178 loaded incidents,
5,196 (3.6\%) fall outside all ZCTA polygons and are dropped.
The remaining 139,982 incidents are assigned to 25 distinct ZCTAs.

\textbf{Temporal binning.}
We discretise time into 1-hour bins, yielding $T = 26{,}280$ hourly
steps over three years. The resulting $(N, T)$ count matrix has
$N = 25$ spatial nodes and is 83.21\% sparse---justifying our choice of
\zinb\ rather than Poisson or Gaussian loss.

\textbf{Feature engineering.}
Each node--timestep pair is described by $F = 13$ features, partitioned into:
\begin{itemize}
  \item \emph{Dynamic count features}: three lagged incident counts
    ($\ell_{t-1}, \ell_{t-2}, \ell_{t-3}$), a 24-hour rolling mean, and a
    7-day rolling mean---all computed with a one-step shift to prevent
    target leakage.
  \item \emph{Periodic time embeddings}: four sine/cosine projections
    encoding hour-of-day and day-of-week cyclically.
  \item \emph{Static ACS demographic features}: percent minority population,
    min-max-normalised median household income, and poverty
    rate~\cite{uscensus2022acs}---broadcast across all timesteps.
  \item \emph{Patrol exposure}: initialised to zero; updated in-place by
    Phase~5 feedback (not used during initial training).
\end{itemize}
The full tensor $\mathbf{X} \in \mathbb{R}^{N \times T \times F}$ is split
chronologically in a 70/15/15 ratio (train/val/test), preserving
time-series integrity without look-ahead.

\subsection{Phase 2: Hybrid Graph Construction}
\label{sec:phase2}

We construct a weighted adjacency matrix $\mathbf{A}_{\text{combined}}
\in \mathbb{R}^{N \times N}$ that combines geographic proximity and
demographic structural similarity:
\begin{equation}
  \mathbf{A}_{\text{combined}} = \alpha_{\text{geo}}\,\mathbf{A}_{\text{geo}}^{\text{norm}}
    + \alpha_{\text{feat}}\,\mathbf{A}_{\text{feat}}^{\theta} + \mathbf{I},
  \label{eq:A_combined}
\end{equation}
where $\alpha_{\text{geo}} = 0.60$, $\alpha_{\text{feat}} = 0.40$,
$\theta = 0.50$ (cosine-similarity threshold), and $\mathbf{I}$
represents self-loops added to the diagonal after combination.

\textbf{Geographic adjacency} $\mathbf{A}_{\text{geo}}$: Queen contiguity,
producing 108 directed edges (mean degree 4.32) with zero isolated nodes.
The matrix is row-normalised to $\mathbf{A}_{\text{geo}}^{\text{norm}}$.

\textbf{Feature-similarity adjacency} $\mathbf{A}_{\text{feat}}$:
Cosine similarity of each ZCTA's ACS feature vector
(percentage minority, normalised median income, poverty rate),
thresholded at 0.50, yielding 484 directed edges (mean degree 19.36).
This captures structural coupling between non-adjacent but demographically
comparable ZCTAs.

The combined adjacency has 513 non-zero entries with edge weights in
$[0.0857, 1.0000]$ (design choice; $N=25$ is small enough for dense
matrix multiplication to be more efficient than sparse operations).

\subsection{Phase 3: Dual-Engine Spatiotemporal Predictor}
\label{sec:phase3}

\subsubsection{Spatiotemporal GNN (\stgnn)}

We adopt a Graph WaveNet-style architecture~\cite{wu2019graphwavenet}
with $L=4$ residual WaveNet blocks.  Each block applies:
(i)~a dilated causal temporal convolution
(dilations $\{1, 2, 4, 8\}$, kernel size 2),
providing a temporal receptive field of 16 hours;
(ii)~a gated tanh-sigmoid activation;
(iii)~graph diffusion via $\mathbf{A}_{\text{combined}}$;
(iv)~skip and residual projections.
The input projection expands $F = 13$ features to $C = 64$ channels;
the output skip accumulator is projected to $D = 32$ embedding dimensions.

\subsubsection{Hawkes Excitation Layer}

Crime events are self-exciting: a cluster of incidents raises the local
probability of subsequent events~\cite{mohler2015randomized,barman2025unmasking}.
We model this with a multivariate Hawkes excitation:
\begin{align}
  \lambda_v(t) &= \mu_v(t) + \mathcal{A}_v(t), \\
  \mathcal{A}_v(t) &= \sum_{k=0}^{H-1} e^{-\beta k} \, z_v(t-1-k), \\
  z_v(t) &= \sum_u \alpha_{uv} \, y_u(t),
\end{align}
where $H = 168$ (7-day history window), and $\alpha_{uv}$, $\beta$ are
learnable parameters initialised at 0.5 and 1.0 respectively via a
\texttt{softplus} reparametrisation.

\subsubsection{ZINB Output Head}

The \stgnn\ embedding $\boldsymbol{\mu}_{\text{embed}}$
(dimension $B \times N \times T \times D$)
is added to the projected Hawkes excitation $\mathcal{A}_{\text{proj}}$,
and the combined representation is decoded by a two-layer MLP into
the three parameters of a Zero-Inflated Negative Binomial distribution:
\begin{align}
  \mu &= \texttt{softplus}(\mathbf{W}_\mu\, h + b_\mu) > 0 \quad \text{(NB mean)}, \\
  \theta &= \texttt{softplus}(\mathbf{W}_\theta\, h + b_\theta) > 0 \quad \text{(dispersion)}, \\
  \pi &= \sigma(\mathbf{W}_\pi\, h + b_\pi) \in (0,1) \quad \text{(zero-inflation gate)}.
\end{align}

The \zinb\ negative log-likelihood is:
\begin{equation}
\begin{split}
  \mathcal{L}_{\text{ZINB}} = -\mathbb{E}_{y}\bigl[
    &\mathbf{1}_{y=0}\log\!\bigl[\pi + (1-\pi)(1+\mu/\theta)^{-\theta}\bigr] \\
    &+\, \mathbf{1}_{y>0}\bigl[\log(1-\pi) + \log p_{\text{NB}}(y;\mu,\theta)\bigr]
  \bigr].
\end{split}
\end{equation}

The full model contains 108,309 trainable parameters.

\subsection{Phase 4: Fairness-Constrained Patrol Allocation}
\label{sec:phase4}

At each timestep $t$, predicted risks $\boldsymbol{\mu}_t \in \mathbb{R}^N$
are passed to a linear programme:
\begin{equation}
  \begin{aligned}
    \max_{\mathbf{p}} \quad & \boldsymbol{\mu}_t^\top \mathbf{p} \\
    \text{s.t.} \quad & \mathbf{1}^\top \mathbf{p} \le B, \\
                      & 0 \le p_v \le B \quad \forall v, \\
                      & \bar{p}_{\text{min}} \ge (1-\varepsilon)\,\bar{p}_{\text{maj}}, \\
                      & \bar{p}_{\text{min}} \le (1+\varepsilon)\,\bar{p}_{\text{maj}},
  \end{aligned}
  \label{eq:alloc_lp}
\end{equation}
where $B = 60$ patrol units, $\varepsilon = 0.05$,
$\bar{p}_{\text{min}}$ is the mean patrol allocation over minority ZCTAs
(percentage minority $\ge 0.5$; 14 of 25 ZCTAs in Baltimore),
and $\bar{p}_{\text{maj}}$ is the mean over non-minority ZCTAs.
The LP is solved per-timestep via CVXPY with warm-start; if infeasible,
the solver falls back to unconstrained proportional allocation.

The Demographic Impact Ratio is $\mathrm{DIR} = \bar{p}_{\text{min}} / \bar{p}_{\text{maj}}$~\cite{chouldechova2017fair,barman2025unmasking}.
A value of $\mathrm{DIR} = 1$ indicates patrol-intensity parity;
the constraint $|\mathrm{DIR}-1| \le 0.05$ limits deviations to 5\%.

\subsection{Phase 5: Deployment Feedback Simulator}
\label{sec:phase5}

We simulate $K = 6$ deployment cycles. Each cycle proceeds in four steps:

\begin{enumerate}
  \item \textbf{Inference.} The current model produces risk estimates
    $\boldsymbol{\mu}_t$ for all $t$ in the test window (3,942 hourly steps).

  \item \textbf{Allocation.} The LP (\Cref{eq:alloc_lp}) produces
    $\mathbf{P} \in \mathbb{R}^{T \times N}$ patrol allocations.

  \item \textbf{Detection simulation.}
    Using a deterministic linear detection model,
    \begin{equation}
      d_v(t) = p_{\text{base}} + (p_{\max} - p_{\text{base}}) \cdot
               \mathrm{clip}\!\left(\frac{P_{t,v}}{B},\, 0, 1\right),
    \end{equation}
    with $p_{\text{base}} = 0.30$ and $p_{\max} = 0.90$,
    the observed count is $\hat{y}_v(t) = y_v(t) \cdot d_v(t)$
    (expected value, not sampled).
    Using expected detection avoids stochastic noise that would mask the
    fairness signal across cycles~\cite{barman2025unmasking}.

  \item \textbf{Incremental retraining.}
    The model is fine-tuned on biased observations $\hat{\mathbf{y}}$
    for 50 epochs using Adam with sliding windows of size 64.
\end{enumerate}

After each cycle, we record: \zinb\ loss, mean DIR and its standard deviation,
Gini concentration coefficient, risk-weighted coverage,
mean minority/majority patrol units, and group-specific detection rates.

\section{Experimental Setup}
\label{sec:setup}

\subsection{Dataset}

We use the Baltimore City Part-1 Crime dataset from the Baltimore Open
Data portal~\cite{baltimore2019data}.
Records are filtered to years 2017, 2018, and 2019, and clipped to a
WGS-84 study bounding box covering latitudes 39.197--39.372 and
longitudes $-$76.713--$-$76.529.
Demographic covariates (percentage minority population, median household
income, and poverty rate) are drawn from the ACS 5-year 2019 ZCTA-level
estimates~\cite{uscensus2022acs}.
Two ZCTAs (21233, 21287) have no ACS coverage and receive
zero-filled demographic features.

\Cref{tab:dataset} summarises the dataset statistics.

\begin{table}[h]
  \centering
  \caption{Baltimore dataset statistics.}
  \label{tab:dataset}
  \small
  \begin{tabular}{lr}
    \toprule
    Statistic & Value \\
    \midrule
    Raw incidents loaded           & 145{,}178        \\
    Incidents dropped (outside ZCTAs) & 5{,}196 (3.6\%) \\
    Incidents assigned to ZCTAs    & 139{,}982        \\
    Spatial nodes (ZCTAs)          & 25               \\
    Temporal bins (1-hour)         & 26{,}280         \\
    Tensor sparsity                & 83.21\%          \\
    Feature dimension $F$          & 13               \\
    Train / Val / Test timesteps   & 18{,}396 / 3{,}942 / 3{,}942 \\
    Minority ZCTAs (pct.\ minority $\ge 0.5$) & 14 / 25 \\
    \bottomrule
  \end{tabular}
\end{table}

\subsection{Graph Configuration}

The hybrid adjacency matrix combines Queen contiguity
($\alpha_{\text{geo}}=0.60$) and cosine feature similarity
($\alpha_{\text{feat}}=0.40$, threshold $\theta=0.50$) with self-loops.
The geographic component produces 108 directed edges (mean degree 4.32);
the feature component produces 484 directed edges (mean degree 19.36);
the combined matrix has 513 non-zero entries with weights in
$[0.0857, 1.0000]$.

\subsection{Model Configuration}

\Cref{tab:hyperparams} lists all model and training hyperparameters.

\begin{table}[h]
  \centering
  \caption{Model and training hyperparameters.}
  \label{tab:hyperparams}
  \small
  \begin{tabular}{ll}
    \toprule
    Hyperparameter & Value \\
    \midrule
    \multicolumn{2}{l}{\textit{STGNN}} \\
    Input channels $F$              & 13 \\
    Residual / skip channels $C$    & 64 \\
    Output embedding $D$            & 32 \\
    Number of WaveNet blocks $L$    & 4  \\
    Temporal kernel size            & 2  \\
    Dilation schedule               & $\{1, 2, 4, 8\}$ \\
    Temporal receptive field        & 16 hours \\
    Dropout                         & 0.20 \\
    \midrule
    \multicolumn{2}{l}{\textit{Hawkes excitation}} \\
    $\alpha$ init (excitation)      & 0.5 \\
    $\beta$ init (decay rate)       & 1.0 \\
    History window $H$              & 168 (7 days) \\
    \midrule
    \multicolumn{2}{l}{\textit{Training}} \\
    Epochs                          & 100       \\
    Batch size (windows)            & 32        \\
    Window size $W$                 & 32        \\
    Learning rate                   & $10^{-3}$ \\
    Weight decay                    & $10^{-4}$ \\
    Gradient clip                   & 5.0       \\
    Random seed                     & 42        \\
    Total parameters                & 108{,}309 \\
    \midrule
    \multicolumn{2}{l}{\textit{Allocation (Phase 4)}} \\
    Patrol budget $B$               & 60 units/timestep \\
    Fairness tolerance $\varepsilon$& 0.05              \\
    Minority threshold              & pct.\ minority $\ge 0.5$ \\
    \midrule
    \multicolumn{2}{l}{\textit{Simulation (Phase 5)}} \\
    Deployment cycles $K$           & 6               \\
    Detection probability range     & $[0.30, 0.90]$  \\
    Retrain epochs per cycle        & 50              \\
    \bottomrule
  \end{tabular}
\end{table}

\subsection{Compute Environment}

Training and simulation were conducted on a single NVIDIA Quadro
RTX~6000 GPU (25.2\,GB VRAM) using CUDA~12.6, cuDNN~91002, and
PyTorch~2.8.0, within a SLURM-managed HPC cluster.
Phase~3 training completed in approximately 53 seconds at roughly
0.20\,s/epoch from epoch~2 onward.
Phase~5 simulation completed in approximately 3 minutes and 31 seconds.

\subsection{Reproducibility}

All code, configuration files, SLURM job scripts, training logs, and
simulation outputs are publicly available in the artifact repository
described in \Cref{sec:artifact}.
The full pipeline can be reproduced by executing the provided job
scripts on a SLURM cluster with a GPU node.
Every numerical result reported in this paper is traceable to a
logged artefact in the repository.

\section{Results}
\label{sec:results}

\subsection{Training and Prediction Performance}
\label{sec:results:training}

\begin{figure}[t]
  \centering
  \begin{subfigure}[b]{0.49\linewidth}
    \includegraphics[width=\linewidth]{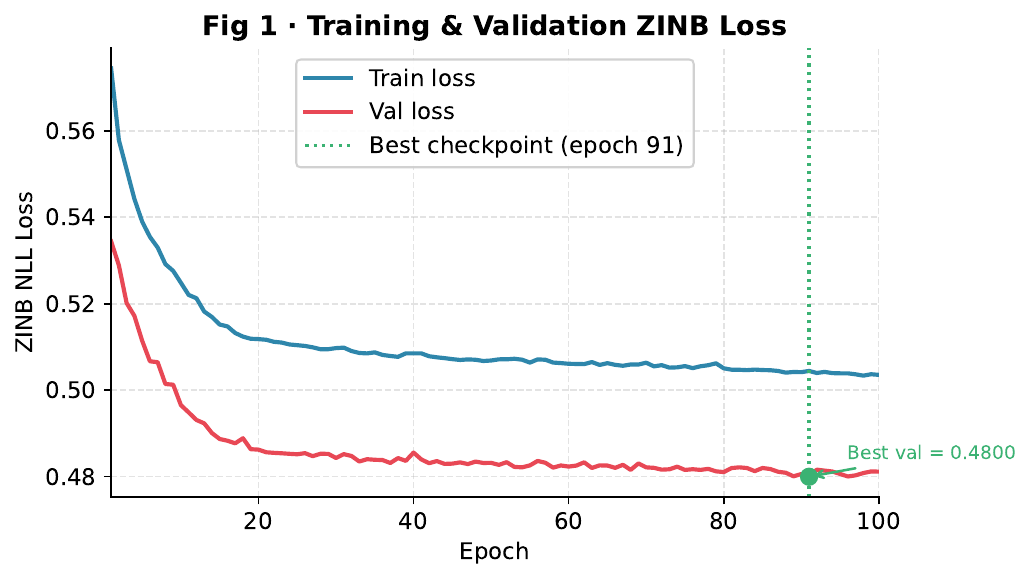}
    \caption{Training and validation \zinb\ loss curves.}
    \label{fig:loss_curve}
  \end{subfigure}
  \hfill
  \begin{subfigure}[b]{0.49\linewidth}
    \includegraphics[width=\linewidth]{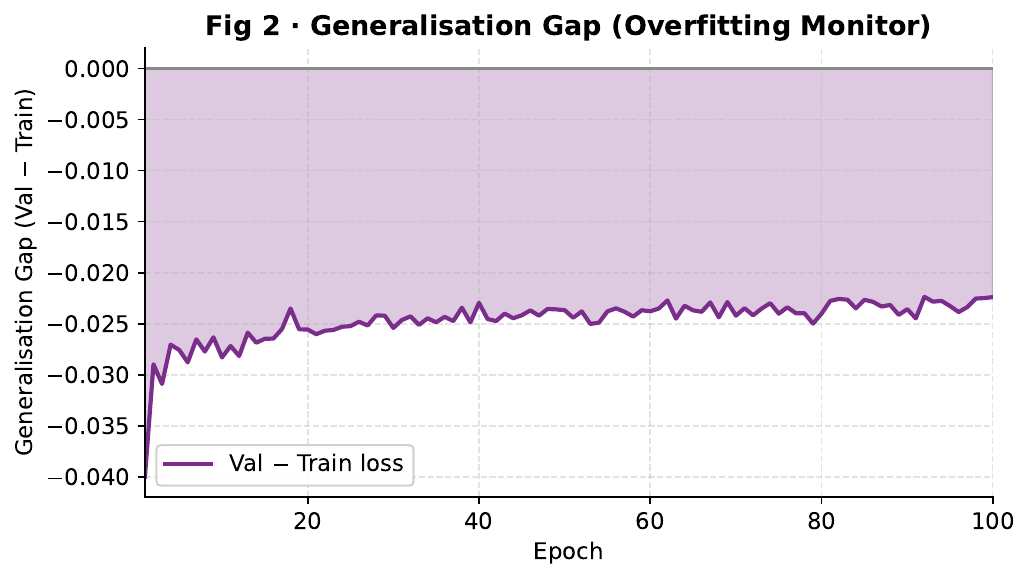}
    \caption{Generalisation gap (Val $-$ Train).}
    \label{fig:loss_gap}
  \end{subfigure}
  \caption{Phase~3 training dynamics over 100 epochs.}
  \label{fig:training}
\end{figure}

\Cref{fig:training} shows the training and validation \zinb\ NLL curves
over 100 epochs.
The model converges smoothly, reaching its best validation loss of
\textbf{0.4800} at epoch~91 (\Cref{fig:loss_curve}).
The test set \zinb\ loss is \textbf{0.4857}, representing a generalisation
gap of 0.0057 relative to the best validation checkpoint.

Training curves show a positive generalisation gap throughout (validation
loss exceeds training loss), consistent with mild overfitting that does
not worsen substantially after epoch~30 (\Cref{fig:loss_gap}).

\textbf{Note on baseline comparison.}
We do not report comparisons against alternative models (\eg\ plain
\stgnn\ without Hawkes, Poisson regression, or DCRNN) because no such
runs were performed under the current codebase.
Ablation experiments are identified as future work
(see \Cref{sec:limitations}).

\subsection{Fairness-Constrained Allocation}
\label{sec:results:allocation}

\begin{figure}[t]
  \centering
  \includegraphics[width=\linewidth]{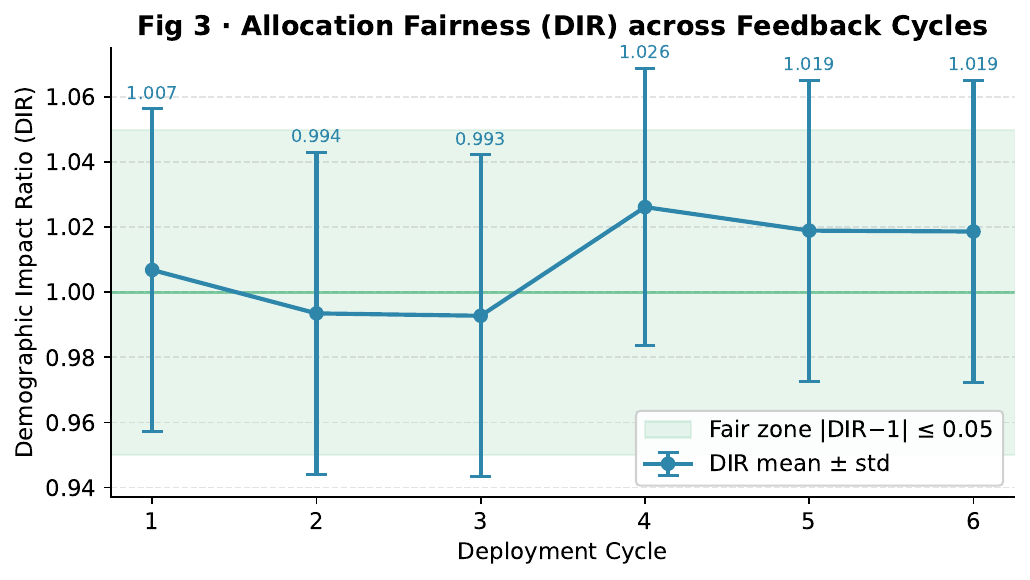}
  \caption{Allocation DIR across 6 deployment cycles with the
    $|\mathrm{DIR}-1|\le 0.05$ fairness band.}
  \label{fig:dir}
\end{figure}

\begin{figure}[t]
  \centering
  \includegraphics[width=\linewidth]{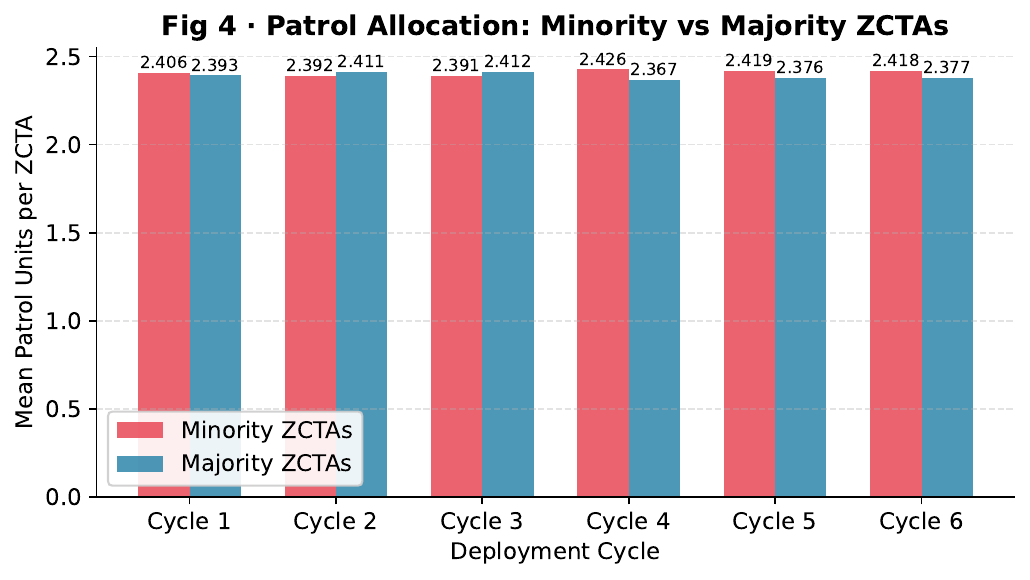}
  \caption{Mean patrol units per ZCTA for minority vs.\ non-minority
    groups across cycles.}
  \label{fig:patrol}
\end{figure}

\Cref{tab:simulation} reports per-cycle metrics from the six-cycle
deployment simulation.
\Cref{fig:dir,fig:patrol,fig:detection,fig:coverage} visualise the
key fairness and allocation trends.

\begin{table*}[t]
  \centering
  \caption{Per-cycle deployment simulation results.
    DIR = Demographic Impact Ratio of patrol allocations.
    Coverage = risk-weighted coverage.
    det\_min / det\_maj = aggregate detection ratio (observed/true) for
    minority / non-minority ZCTAs.}
  \label{tab:simulation}
  \small
  \begin{tabular}{ccccccc}
    \toprule
    Cycle & ZINB Loss & DIR (mean $\pm$ std) & Gini & Coverage
          & det\_min & det\_maj \\
    \midrule
    1 & 0.2173 & $1.0068 \pm 0.0495$ & 0.9249 & 0.9359 & 0.3391 & 0.3785 \\
    2 & 0.2155 & $0.9935 \pm 0.0496$ & 0.9246 & 0.8763 & 0.3411 & 0.3783 \\
    3 & 0.2198 & $0.9928 \pm 0.0495$ & 0.9246 & 0.8858 & 0.3422 & 0.3791 \\
    4 & 0.2158 & $1.0262 \pm 0.0426$ & 0.9253 & 0.8827 & 0.3429 & 0.3777 \\
    5 & 0.2157 & $1.0189 \pm 0.0463$ & 0.9251 & 0.9010 & 0.3428 & 0.3780 \\
    6 & 0.2154 & $1.0187 \pm 0.0464$ & 0.9251 & 0.9016 & 0.3429 & 0.3780 \\
    \bottomrule
  \end{tabular}
\end{table*}

\subsubsection{Allocation Fairness (DIR)}

The mean patrol-allocation DIR across cycles ranges from
0.9928 (cycle~3) to 1.0262 (cycle~4), well within the
$|\mathrm{DIR}-1| \le 0.05$ constraint band (\Cref{fig:dir}).
The per-timestep DIR standard deviation ($0.043$--$0.050$)
reflects natural variation in the allocation as the risk surface changes
across the 3,942 timesteps in the test window.
These results confirm that the LP constraint is satisfied on average;
individual timestep violations are possible (the constraint applies to
the expected mean, not each step independently), but the aggregate DIR
remains within tolerance across all six cycles.

\subsubsection{Patrol Distribution}

\Cref{fig:patrol} shows that mean patrol allocation per ZCTA remains
close to parity between minority ZCTAs ($2.39$--$2.43$ units/timestep)
and non-minority ZCTAs ($2.37$--$2.41$ units/timestep) across all cycles.
These values imply a total per-cycle patrol deployment of
approximately $25 \times 2.4 = 60$ units---consistent with the
configured budget $B = 60$.

\subsubsection{Coverage and Concentration}

\begin{figure}[t]
  \centering
  \includegraphics[width=\linewidth]{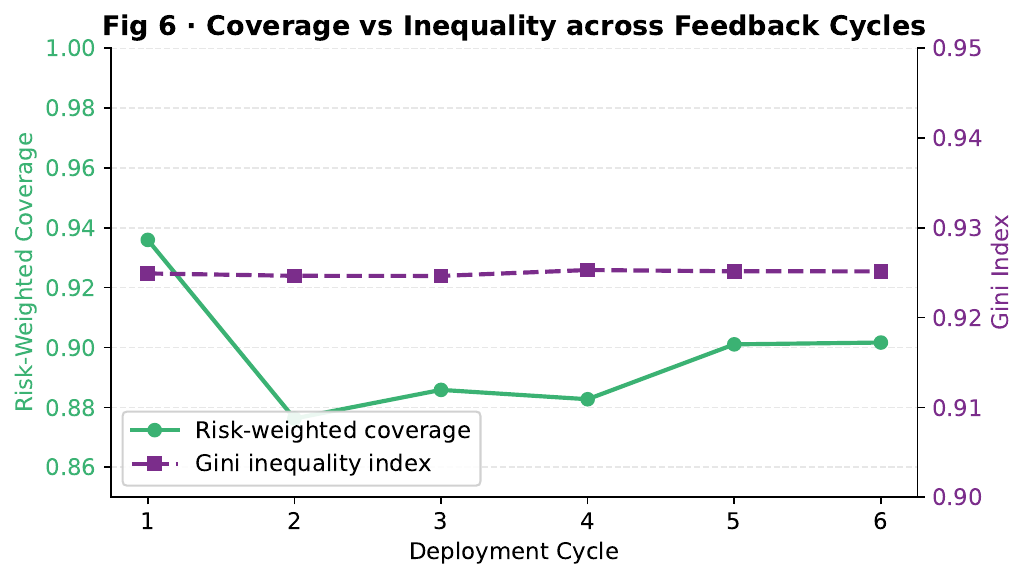}
  \caption{Risk-weighted coverage and Gini index across cycles.}
  \label{fig:coverage}
\end{figure}

Risk-weighted coverage---the fraction of the total predicted risk
mass that falls within patrolled ZCTAs---ranges from 0.876 (cycle~2)
to 0.936 (cycle~1), with cycles 5 and 6 stabilising around 0.901
(\Cref{fig:coverage}).
The Gini coefficient of the allocation is consistently high
($\approx 0.925$), indicating that patrol is concentrated in a
small fraction of ZCTAs at any given timestep---a consequence of
optimising for risk-weighted coverage with a limited budget.

\subsection{Detection Rate Disparity and Feedback Bias}
\label{sec:results:detection}

\begin{figure}[t]
  \centering
  \includegraphics[width=\linewidth]{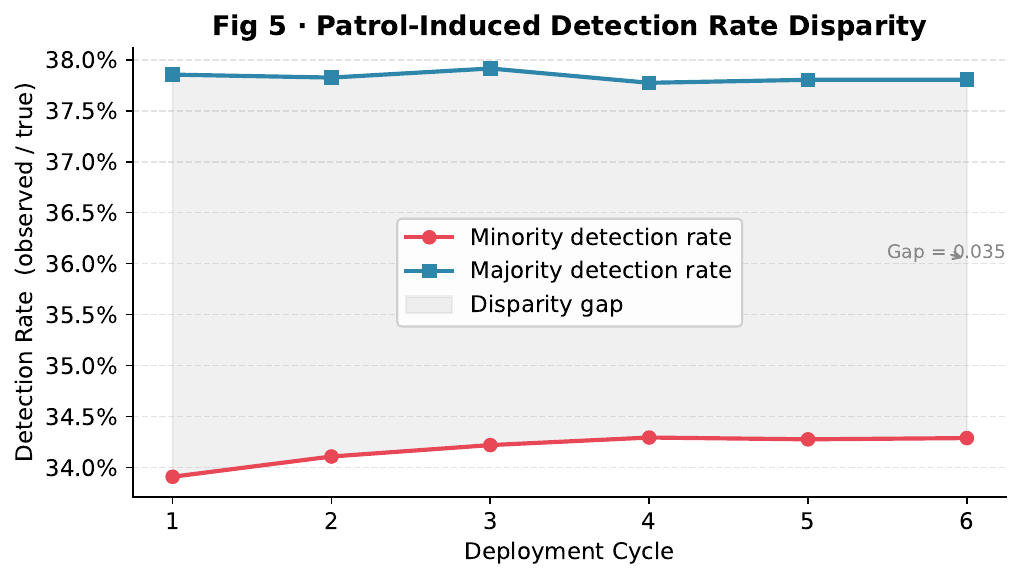}
  \caption{Aggregate detection rates (observed / true counts)
    for minority vs.\ non-minority ZCTAs across cycles.}
  \label{fig:detection}
\end{figure}

\Cref{fig:detection} shows the aggregate detection ratios---the ratio of
total observed to total true incident counts---for minority ZCTAs
(0.339--0.343) and non-minority ZCTAs (0.378--0.379) across all six
cycles.  The detection rate is systematically higher in non-minority ZCTAs
by approximately 3.5 percentage points throughout the simulation.

This disparity is a direct consequence of the allocation LP's interaction
with the linear detection model: even near-parity patrol allocation (DIR
$\approx 1$) results in higher effective detection in non-minority ZCTAs
because those ZCTAs receive proportionally more patrol relative to their
true crime burden (they have fewer incidents on average).
This is a form of feedback-induced observational bias that the
allocation-level DIR constraint does not fully correct, because DIR
measures patrol-unit parity, not detection-outcome
parity~\cite{ensign2018runaway,chouldechova2017fair,barman2025unmasking}.

\begin{figure}[t]
  \centering
  \includegraphics[width=\linewidth]{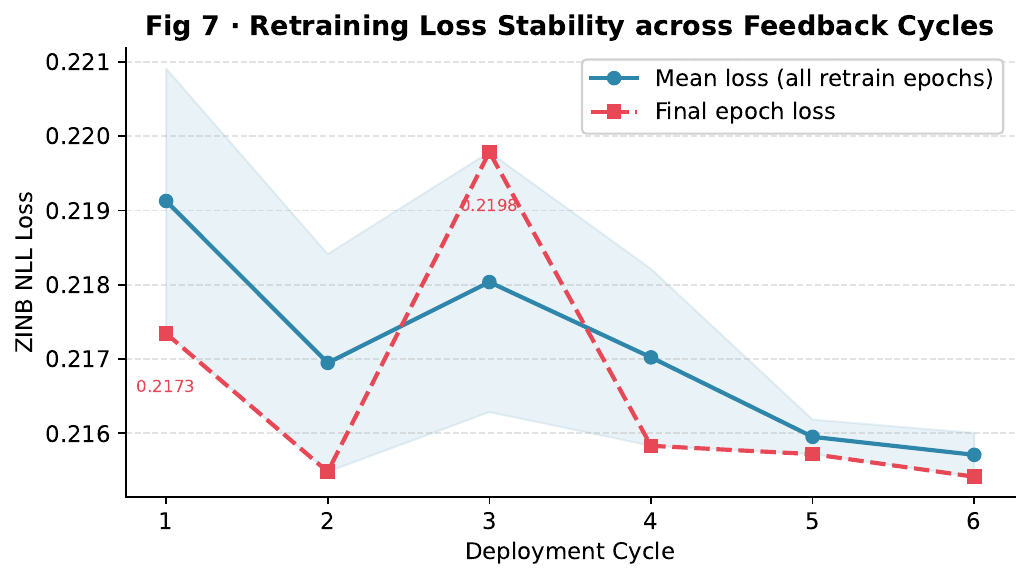}
  \caption{\zinb\ retraining loss per cycle (mean and final epoch).}
  \label{fig:retrain}
\end{figure}

The \zinb\ retraining loss across cycles is stable, ranging from
0.2154 to 0.2198 (\Cref{fig:retrain}), indicating that the model
adapts to the biased observation stream without catastrophic divergence.

\textbf{Note on observed-count DIR values.}
The simulation also records an observed-count DIR (ratio of mean
observed incidents in minority vs.\ non-minority ZCTAs),
which reaches values on the order of $24{,}000$--$24{,}700$ across
cycles.  These extreme values arise from additive smoothing applied to
the denominator when non-minority ZCTAs have near-zero observed counts
across many timesteps.
We do not interpret these values as meaningful fairness measurements
and exclude them from the main analysis.

\section{Limitations}
\label{sec:limitations}

We identify the following limitations explicitly.
Omitting these would overstate the maturity and generalisability of \fase.

\textbf{Single city, single study period.}
All experiments use Baltimore Part-1 crime data for 2017--2019.
No transfer learning, cross-city validation, or temporal
out-of-distribution evaluation has been performed.
Generalisation to other cities, crime types, or time windows is unknown.

\textbf{No baseline comparisons.}
The reported \zinb\ losses (val 0.4800, test 0.4857) are absolute values
of a loss function with no universally agreed-upon scale.
We did not run ablation experiments (e.g., \stgnn\ without Hawkes,
Poisson regression, or existing competitive models such as
DCRNN~\cite{li2018dcrnn} or fairness-aware crime prediction
models~\cite{wu2024fairness}).
The absence of baselines means we cannot claim predictive superiority
over simpler approaches.

\textbf{Fairness metric limitations.}
The DIR constraint measures patrol-intensity parity, not outcome parity.
As demonstrated in \Cref{sec:results:detection}, allocation-level DIR
$\approx 1$ coexists with a persistent detection-rate gap of approximately
3.5 percentage points between minority and non-minority ZCTAs.
Alternative fairness criteria (equal opportunity, calibration, or
outcome-based constraints) may yield different conclusions.
Counterfactual and causal fairness formulations~\cite{kim2021counterfactual,zhang2018fairness}
offer theoretically stronger guarantees but require individual-level
causal models that are not available in aggregate crime-count settings.
In particular, individual-level or intersectional fairness is not addressed.

\textbf{Simulated, not real, deployment.}
Phase~5 uses a deterministic linear detection model ($p_{\text{base}}=0.30$,
$p_{\max}=0.90$) to generate observed counts.
Real-world detection probability is heterogeneous across crime types,
reporting cultures, and officer behaviour, and is unlikely to follow
a simple linear functional form~\cite{lum2016predict,barman2025unmasking}.
The feedback simulator should be viewed as a controlled experimental
device for measuring the direction and rough magnitude of feedback effects,
not as an accurate representation of real patrol-crime dynamics.

\textbf{Static graph, no temporal graph evolution.}
The ZCTA adjacency graph is fixed throughout training and simulation.
In practice, spatial relationships and demographic compositions change
over time.

\textbf{Missing ACS coverage.}
Two ZCTAs (21233 and 21287) receive zero-filled demographic features
due to absent ACS data~\cite{uscensus2022acs}.
These ZCTAs are included in the graph and allocation but their demographic
features are uninformative.

\textbf{High Gini coefficient.}
The patrol Gini index ($\approx 0.925$) indicates extreme concentration
of patrol resources.  Whether this is desirable or harmful depends on
context and values~\cite{selbst2019fairness,berk2021fairness}; the
system optimises risk-weighted coverage, not geographic equity.

\textbf{Generalization gap stability.}
The generalisation gap (val $-$ train \zinb\ loss) is positive throughout
training but does not diverge, suggesting mild overfitting.
More aggressive regularisation or early stopping with patience might
further reduce the test loss.

\section{Conclusion}
\label{sec:conclusion}

We presented \fase, a five-phase research pipeline that unifies
spatiotemporal crime prediction, fairness-constrained patrol allocation,
and closed-loop deployment feedback simulation.
Applied to 139,982 Baltimore Part-1 crime incidents (2017--2019)
discretised into 26,280 one-hour bins across 25 ZCTAs,
the dual-engine \stgnn{+}Hawkes predictor achieves a \zinb\ validation
loss of 0.4800 and a test loss of 0.4857.

Fairness-constrained linear programming allocates 60 patrol units per
timestep with a Demographic Impact Ratio (DIR) within
$[0.9928, 1.0262]$ across six deployment cycles---satisfying the
$\varepsilon=0.05$ DIR constraint in aggregate.
Risk-weighted coverage stabilises at approximately 0.901 in cycles 5--6.

A key finding is the persistence of a detection-rate disparity despite
allocation-level fairness: minority ZCTAs consistently exhibit a 3.5
percentage-point lower aggregate detection rate than non-minority ZCTAs
across all six cycles.
This demonstrates that allocation fairness, as measured by DIR, does not
imply equal observational quality in the retraining data---a distinction
that matters for any system that learns from police-reported
incidents~\cite{barman2025unmasking,semsar2026comparative}.

\fase\ is fully open-source and reproducible, with all results verifiable
from logged artefacts.
Future work should include (i)~comparative baselines against models
such as DCRNN~\cite{li2018dcrnn} and fairness-aware
predictors~\cite{wu2024fairness,wang2023pursuit},
(ii)~ablation of the Hawkes component,
(iii)~outcome-parity and causal fairness
criteria~\cite{kim2021counterfactual,zhang2018fairness},
(iv)~multi-city generalisation extending our prior
work~\cite{barman2025unmasking}, and
(v)~a more realistic detection model derived from empirical
clearance-rate data.

\section{Artifact Availability}
\label{sec:artifact}

The \fase\ codebase, configuration files, training logs, simulation
results, and paper figures are publicly available at:
\begin{center}
  \url{https://github.com/pronob29/fase-predictive-policing-framework}
\end{center}

\subsection{Reproducibility}

\textbf{One-command pipeline.}
Running the provided top-level SLURM job script queues all pipeline
phases as sequential dependent jobs on a GPU node.

\textbf{Phase-by-phase reproduction.}
Individual phases can be re-run using the corresponding SLURM scripts
in the \texttt{jobs/} directory.
Each script specifies its required inputs and produced outputs;
full details are given in the repository \texttt{README}.

\begin{itemize}
  \item \textbf{Phase 1+2} (data and graph construction):
    requires the raw Baltimore crime CSV, ACS demographic CSV,
    and Census ZCTA shapefile; produces processed tensors and
    the pre-built graph.

  \item \textbf{Phase 3} (model training):
    requires Phase 1+2 outputs and a CUDA-capable GPU;
    produces the best model checkpoint and training log.

  \item \textbf{Phase 5} (feedback simulation):
    requires the Phase 3 checkpoint;
    produces the per-cycle simulation history used in all results tables.
\end{itemize}

\textbf{Figure generation.}
All seven paper figures can be regenerated from the simulation
history and training log using the provided visualisation module
(\texttt{python -m fase.visualize}); see the repository
\texttt{README} for the full invocation.

\begin{sloppypar}
\textbf{Software dependencies.}
Core libraries (see \texttt{requirements.txt}): Python~3.9+,
PyTorch~2.x, PyTorch~Geometric, \texttt{cvxpy},
\texttt{geopandas}, \texttt{libpysal}, \texttt{pandas},
\texttt{numpy}, \texttt{matplotlib}.
\end{sloppypar}

\subsection{Data Availability}

The raw Baltimore crime data is available at the Baltimore Open Data
portal (\url{https://data.baltimorecity.gov}).
ACS 5-year (2019) ZCTA-level estimates are available from the US Census
Bureau (\url{https://data.census.gov}).
ZCTA shapefiles are available from the US Census TIGER/Line programme.
Raw input files are not redistributed in the repository due to licensing;
the repository \texttt{README} includes download instructions.

\bibliographystyle{ACM-Reference-Format}
\bibliography{references}

\end{document}